\documentclass[conference]{IEEEtran}
\IEEEoverridecommandlockouts

\usepackage{cite}
\usepackage{amsmath,amssymb,amsfonts}
\usepackage{graphicx}
\usepackage{textcomp}
\usepackage{xcolor}
\usepackage{booktabs}
\usepackage{multirow}
\usepackage{url}
\usepackage{algorithm}
\usepackage{algpseudocode}
\usepackage{fancyhdr}
\usepackage{colortbl}
\usepackage{caption}
\usepackage{adjustbox}
\usepackage{hyperref}
\hypersetup{
    colorlinks=true,
    citecolor=blue,
    linkcolor=black,
    urlcolor=blue
}

\algrenewcommand\alglinenumber[1]{\small\mdseries #1:}

\def\BibTeX{{\rm B\kern-.05em{\sc i\kern-.025em b}\kern-.08em
    T\kern-.1667em\lower.7ex\hbox{E}\kern-.125emX}}

\begin{document}

\title{VLAgeBench: Benchmarking Large Vision-Language Models for Zero-Shot Human Age Estimation\\
}

\author{
    \IEEEauthorblockN{
        Rakib Hossain Sajib\textsuperscript{1}, 
        Md Kishor Morol\textsuperscript{2}, 
        Rajan Das Gupta\textsuperscript{3}, 
        Mohammad Sakib Mahmood\textsuperscript{4},\\
        Shuvra Smaran Das\textsuperscript{5}
    } 
    \IEEEauthorblockA{
        \textsuperscript{1}Department of Computer Science and Engineering, Begum Rokeya University, Rangpur, Rangpur, Bangladesh  \\
        \textsuperscript{2,4,5}EliteLab.AI, Queens, New York, United States \\
        \textsuperscript{3}Department of Computer Science, American International University - Bangladesh, Dhaka, Bangladesh\\
        Email: rakibnsajib@gmail.com,
        kishor@elitelab.ai, 
        18-36304-1@student.aiub.edu,\\
        sakib.mahmood@elitelab.ai,
        shuvra.das@elitelab.ai
    }
}

\maketitle
\thispagestyle{fancy} 
\fancyhf{} 

\renewcommand{\headrulewidth}{0pt}
\renewcommand{\footrulewidth}{0pt}

\begin{abstract}
Human age estimation from facial images represents a challenging computer vision task with significant applications in biometrics, healthcare, and human-computer interaction. While traditional deep learning approaches require extensive labeled datasets and domain-specific training, recent advances in large vision-language models (LVLMs) offer the potential for zero-shot age estimation. This study presents a comprehensive zero-shot evaluation of state-of-the-art Large Vision-Language Models (LVLMs) for facial age estimation, a task traditionally dominated by domain-specific convolutional networks and supervised learning. We assess the performance of GPT-4o, Claude 3.5 Sonnet, and LLaMA 3.2 Vision on two benchmark datasets, UTKFace and FG-NET, without any fine-tuning or task-specific adaptation. Using eight evaluation metrics, including MAE, MSE, RMSE, MAPE, MBE, $R^2$, CCC, and $\pm$5-year accuracy, we demonstrate that general-purpose LVLMs can deliver competitive performance in zero-shot settings. Our findings highlight the emergent capabilities of LVLMs for accurate biometric age estimation and position these models as promising tools for real-world applications. Additionally, we highlight performance disparities linked to image quality and demographic subgroups, underscoring the need for fairness-aware multimodal inference. This work introduces a reproducible benchmark and positions LVLMs as promising tools for real-world applications in forensic science, healthcare monitoring, and human-computer interaction. The benchmark focuses on strict zero-shot inference without fine-tuning and highlights remaining challenges related to prompt sensitivity, interpretability, computational cost, and demographic fairness.
\end{abstract}


\section{Introduction}

Estimating a person’s age from a facial image is a fundamental problem in computer vision, with broad applications in healthcare, security, social media moderation, and human-computer interaction. Traditional approaches rely heavily on deep learning models such as convolutional neural networks (CNNs), trained on labeled datasets like UTKFace and FG-NET \cite{rothe2015dex, panwar2021survey}. These models learn patterns from facial features to predict age accurately when the test data closely resembles the training data. However, they often struggle with generalization, particularly across variations in ethnicity, lighting conditions, facial expressions, and other real-world diversity \cite{agbo2021facial, 9736350, 10903352}. Recently, Large Vision-Language Models (LVLMs) such as GPT-4o, Claude 3.5 Sonnet, and LLaMA 3.2 Vision have shown strong multimodal capabilities across tasks like captioning and classification without fine-tuning \cite{xu2024exploring}. However, their performance on precise, regression-based tasks like facial age estimation remains largely unexplored. CLIP-based zero-shot approaches to ordinal regression have demonstrated modest performance in similar settings, often reporting higher MAEs compared to trained CNNs \cite{ordiclIP}. Facial age prediction is challenging as it requires precise regression outputs based on subtle cues like wrinkles and skin texture. Zero-shot performance is highly sensitive to prompt design \cite{wang2023prompt}, and fairness concerns arise due to potential biases inherited from training data, affecting demographic consistency \cite{panic2024bias}.
\par In this study, we conduct the first comprehensive zero-shot evaluation of LVLMs for facial age estimation. We benchmark three leading large vision-language models such as GPT-4o, Claude 3.5 Sonnet and LLaMA 3.2 Vision on the UTKFace and FG-NET datasets, which include individuals from diverse age ranges, genders, and ethnic backgrounds \cite{xu2024exploring}. Unlike previous studies focused on classification tasks, our work explores whether LVLMs can infer continuous age values from facial images in a zero-shot setting, without task-specific training. Using a standardized prompt across all models, we evaluate their outputs using multiple regression-based metrics, including Mean Absolute Error (MAE), Root Mean Squared Error (RMSE), R\textsuperscript{2}, Concordance Correlation Coefficient (CCC), and accuracy within $\pm$5 years.

\vspace{1mm}
\par Our results show that GPT-4o achieves the best performance, with MAEs of 4.93 on UTKFace and 3.73 on FG-NET, and high accuracy within $\pm$5 years. However, all models face issues like prompt sensitivity and demographic bias. This work presents a new zero-shot benchmark for facial age estimation with VLMs and provides insights to improve fairness, robustness, and generalizability in multimodal AI systems. Unlike fine-tuned or few-shot LVLM approaches, our study intentionally focuses on pure zero-shot generalization to establish a training-free baseline for age estimation.

The core contributions of this work are:
\begin{itemize}
    \item Presenting the first systematic benchmark evaluation of state-of-the-art LVLMs for zero-shot facial age estimation, establishing baseline performance metrics across standardized datasets.
    
    \item Conducting a comprehensive performance analysis using eight complementary evaluation metrics that capture both prediction accuracy and agreement between estimated and ground-truth ages.
    
    \item Analyzing demographic fairness and model robustness across diverse population subgroups, highlighting critical considerations for real-world deployment.
    
    \item Providing reproducible experimental protocols and open-source implementation to facilitate future research in this emerging area.
\end{itemize}


\section{Literature Review}

Recent advances in Vision-Language Models (VLMs) have transformed multimodal learning by enabling generalizable image-text understanding through large-scale pretraining and zero-shot inference. In contrast, traditional age estimation relies on supervised CNNs (e.g., VGG, ResNet) and ViTs trained on datasets such as FG-NET, UTKFace, and IMDB-WIKI, but these models struggle under demographic shifts \cite{Dahlan2021}. Georgopoulos et al.~\cite{Georgopoulos2018} highlighted the role of handcrafted morphological and kinship-based features, supported by data augmentation, in improving age prediction.

\par The vision-language paradigm began with models like CLIP and ALIGN, which align visual and textual embeddings using contrastive learning, showing strong zero-shot classification performance \cite{Zhang2024}. However, their capacity for fine-grained regression tasks, such as age estimation, remains limited. Recent models such as BLIP, Flamingo, MiniGPT-4, and GPT-4o build upon this foundation by employing unified or dual-stream transformer architectures and generative learning objectives \cite{Chen2023}. GPT-4o, in particular, enables direct image-text interaction and produces free-form outputs, including accurate numerical estimates \cite{10681094}. These models show promise in attribute recognition tasks. AlDahoul et al.~\cite{AlDahoul2024} evaluated Claude, GPT-4o, and PaliGemma on facial attribute recognition, including age group classification, and found that fine-tuned VLMs could outperform conventional CNNs.

\par Prompt formulation critically affects zero-shot performance. Zhou et al.~\cite{Zhou2022} showed that models like CLIP are highly sensitive to prompt wording and proposed learnable context prompts (CoOp) to improve consistency. Subsequent work by Zhang et al.~\cite{Zhang2024b} introduced mutual-information-based prompt alignment, and Cho et al.~\cite{Cho2023} proposed distribution-aware prompt tuning to reduce domain shift during inference. PromptMargin~\cite{Anonymous2024} jointly optimizes visual and textual prompts to enhance discrimination in low-data scenarios.

\par Despite their power, VLMs exhibit demographic bias and hallucination tendencies. Bordes et al.~\cite{Bordes2023} and Liang et al.~\cite{Liang2024} highlight that VLMs often conflate visual semantics with cultural or contextual artifacts, raising fairness concerns especially in biometric tasks like age estimation. Luo et al.~\cite{Luo2024} introduced FairCLIP, a dataset with explicit demographic labels revealing significant accuracy disparities across gender and ethnicity groups, even for well-trained models like BLIP and CLIP. Our use of UTKFace and FG-NET aims to address these concerns by benchmarking performance across diverse demographics.

\par Although age estimation is underexplored in VLM research, interest is growing. AlDahoul et al.~\cite{AlDahoul2024} demonstrated GPT-4o’s zero-shot ability for age estimation with some difficulties on extreme age groups. Zeng et al.~\cite{Zeng2024} applied GPT-4 to zero-shot building age estimation from facade images, showing the approach’s generality.

\par To our knowledge, this work is the first to systematically benchmark LVLMs for direct numeric age regression in a zero-shot setting, using multiple regression and agreement metrics.

\section{Methodology}

\begin{figure*}
    \centering
    \includegraphics[width=1\linewidth]{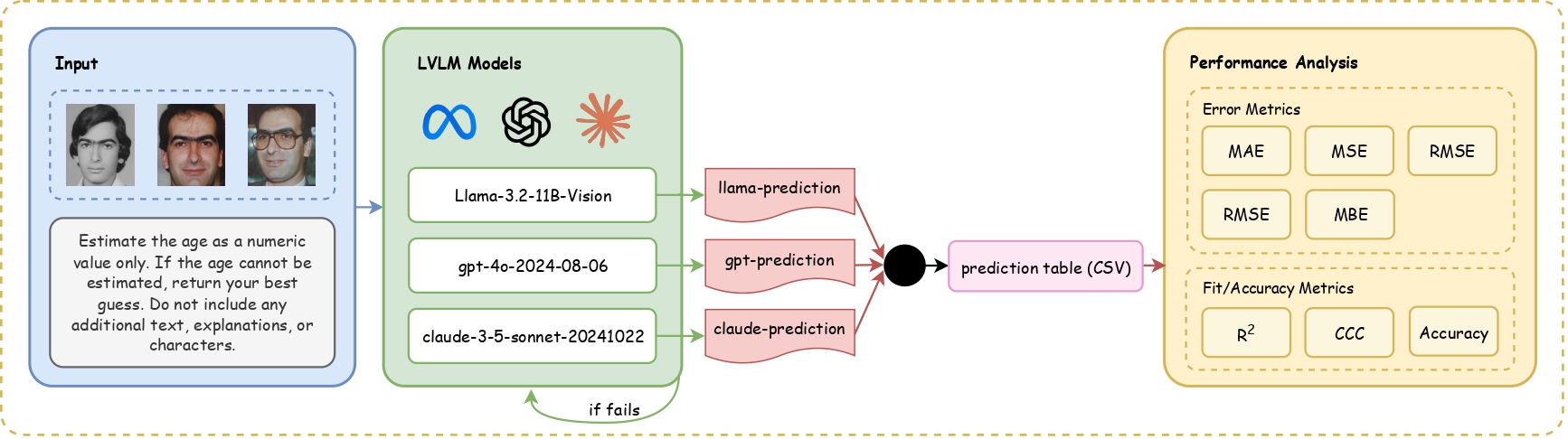}
    \caption{Pipeline for Age Estimation Using Large Vision-Language Models (LVLMs). Input images are processed by three LVLMs (LLaMA 3.2 Vision, GPT-4o, and Claude 3.5 Sonnet) to generate individual age predictions, which are then collected in a unified prediction table (CSV) and evaluated using identical performance metrics.}
    \label{fig:vllm-pipeline}
\end{figure*}

\subsection{Datasets}

This study utilizes two main datasets: UTKFace~\cite{Zhang_2017_CVPR} and FG-NET~\cite{7159107}. The UTKFace dataset includes over 20,000 images that are annotated with information regarding age, gender, and ethnicity. The dataset covers ages from 0 to 116 years, ensuring a thorough basis for model evaluation. The file name of each face image contains embedded labels, structured as [age]\_[gender]\_[race]\_[date\&time].jpg. The dataset's broad age range and demographic diversity make it particularly suitable for evaluating model generalization across population subgroups.

The FG-NET dataset includes a total of 1,002 images collected from 82 individuals, spanning ages from 0 to 69 years. The labels follow the structure [id]A[age].JPG, with id representing the subject identifier and age indicating the annotated age. The historical characteristics of the dataset introduce challenges, including low resolution and inconsistent lighting.

\subsection{Models}

The evaluation of the model was conducted in a zero-shot manner utilizing  large vision-language models (LVLMs). We incorporate both the commercial state-of-the-art model GPT-4o (gpt-4o-2024-08-06), Claude 3.5 Sonnet (claude-3-5-sonnet-20241022), and the open-source model LLaMA 3.2 Vision (LLaMA-3.2-11B-Vision-Instruct).

\subsection{Evaluation Metrics}

To evaluate the performance of the models, we used several metrics commonly employed in regression tasks. The equations for these metrics are as follows:

\subsubsection{Mean Absolute Error (MAE)}
\begin{equation}
\text{MAE} = \frac{1}{N}\sum_{i=1}^{N}|y_i - \hat{y}_i|
\end{equation}
MAE \cite{Reaj2025xFEBERT} measures the average magnitude of absolute errors between predicted and actual ages.

\subsubsection{Mean Squared Error (MSE)}
\begin{equation}
\text{MSE} = \frac{1}{N}\sum_{i=1}^{N}(y_i - \hat{y}_i)^2
\end{equation}
MSE \cite{haque2021erp} penalizes larger errors more significantly compared to MAE.

\subsubsection{Root Mean Squared Error (RMSE)}
\begin{equation}
\text{RMSE} = \sqrt{\text{MSE}}
\end{equation}
RMSE \cite{taslimul2026role} represents the square root of MSE, giving errors in the same units as the predictions.

\subsubsection{Mean Absolute Percentage Error (MAPE)}
\begin{equation}
\text{MAPE} = \frac{1}{N}\sum_{i=1}^{N}\frac{|y_i - \hat{y}_i|}{y_i} \times 100
\end{equation}
MAPE \cite{taslimul2026erm} expresses error as a percentage of actual values, useful for relative comparisons.

\subsubsection{Mean Bias Error (MBE)}
\begin{equation}
\text{MBE} = \frac{1}{N}\sum_{i=1}^{N}(\hat{y}_i - y_i)
\end{equation}
MBE indicates whether the model tends to overestimate or underestimate age.

\subsubsection{Coefficient of Determination ($R^2$)}
\begin{equation}
R^2 = 1 - \frac{\sum_{i=1}^{N}(y_i - \hat{y}_i)^2}{\sum_{i=1}^{N}(y_i - \bar{y})^2}
\end{equation}
$R^2$ \cite{jawadul2026health} measures how well the predictions explain the variance in the actual values.

\subsubsection{Concordance Correlation Coefficient (CCC)}
\begin{equation}
\text{CCC} = \frac{2\rho\sigma_y\sigma_{\hat{y}}}{\sigma_y^2 + \sigma_{\hat{y}}^2 + (\mu_y - \mu_{\hat{y}})^2}
\end{equation}
CCC combines measures of precision and accuracy to evaluate prediction agreement.

\subsubsection{Accuracy within ±5 Years}
\begin{equation}
\text{Accuracy}_{\pm 5\text{years}} = \frac{\# \text{predictions with } |y_i - \hat{y}_i| \leq 5}{N} \times 100
\end{equation}
This metric calculates the percentage of predictions where the error is within ±5 years of the actual age.

\begin{algorithm}[t]
  \caption{Age Estimation using LVLM}
  \label{alg:age-estimation}
  \begin{algorithmic}[1]
    \State \textbf{def} {\large L}OAD\_\kern0pt{\large I}MAGES():
    \Statex \quad Images = Load\_Split\_Dataset() \(\triangleright\) Load images from dataset split
    \State \quad \textbf{return} Images

    \State \textbf{def} {\large G}ENERATE\_\kern0pt{\large P}ROMPT(I):
    \Statex \quad \(\tilde P\) = ``Estimate the age as a numeric value only. If the age cannot be estimated, return your best guess. Do not include any additional text, explanations, or characters."
    \State \quad \textbf{return} \(\tilde P\)

    \State \textbf{def} {\large E}STIMATE\_\kern0pt{\large A}GE(I,\(\tilde P\)):
    \Statex \quad \(\hat A\) = LVLM(\(\tilde P\), I) \(\triangleright\) Pass image and prompt to LVLM
    \State \quad \textbf{return} \(\hat A\)

    \State \textbf{def} {\large A}GE\_\kern0pt{\large E}STIMATION():
    \Statex \quad Images = LOAD\_\kern0ptIMAGES()
    \Statex \quad Predicted\_Ages = []
    \Statex \quad \textbf{for} each I in Images:
    \Statex \quad\quad \(\tilde P\) = GENERATE\_\kern0ptPROMPT(I)
    \Statex \quad\quad \(\hat A\) = ESTIMATE\_\kern0pt AGE(I,\(\tilde P\))
    \Statex \quad\quad Predicted\_Ages.append(\(\hat A\))
    \State \quad \textbf{return} Predicted\_Ages
  \end{algorithmic}
\end{algorithm}

\subsection{Experimental Setup}

The open-source LLaMA 3.2 Vision model had been run on an NVIDIA A100 GPU with 40GB of VRAM to facilitate efficient inference. The commercial models, including GPT-4o and Claude 3.5 Sonnet, were accessed using their respective APIs. We note that such LVLM-based inference is computationally more expensive than traditional CNN-based age estimators, which may limit large-scale or real-time deployment. To ensure fairness and comparability, all models were assessed under standardized conditions with uniform input preprocessing, consistent evaluation metrics, and identical task settings, thereby reducing bias and facilitating a rigorous evaluation of their relative performance in the zero-shot age estimation task utilizing LVLMs.

To elucidate our methodology, we illustrate the comprehensive pipeline in Fig.~\ref{fig:vllm-pipeline}. This figure presents an extensive overview of the process involved in age estimation utilizing large vision-language models (LVLMs). Each LVLM is evaluated independently using identical regression metrics; no ensemble fusion of predictions is performed.

\section{Result Analysis}
The performance of LVLMs has been evaluated on two benchmark age estimation datasets. Multiple error- and correlation-based metrics are considered, with both dataset-specific outcomes and cross-dataset patterns examined.

\begin{table*}[htbp]
\caption{Comprehensive Performance Evaluation of LVLMs for Zero-Shot Age Estimation}
\begin{center}
\begin{tabular}{|c|c|c|c|c|c|c|c|c|c|}
\hline
\textbf{Dataset} & \textbf{Model} & \textbf{MAE} & \textbf{MSE} & \textbf{RMSE} & \textbf{MAPE} & \textbf{MBE} & $\mathbf{R^2}$ & \textbf{CCC} & \textbf{ACC. ±5yrs(\%)} \\
\hline
\multirow{3}{*}{UTKFace} 
& LLaMA 3.2 Vision & 7.58 & 109.82 & 10.48 & 67.40 & 0.01 & 0.72 & 0.86 & 51.48 \\
\cline{2-10}
& Claude 3.5 Sonnet & 5.28 & 55.13 & 7.43 & 30.28 & 0.34 & 0.86 & 0.93 & 64.14 \\
\cline{2-10}
& GPT-4o & 4.93 & 47.35 & 6.88 & 18.48 & -1.36 & 0.88 & 0.94 & 66.42 \\
\hline
\multirow{3}{*}{FGNET} 
& LLaMA 3.2 Vision & 4.86 & 47.34 & 6.88 & 2.22×10$^9$ & -3.66 & 0.71 & 0.87 & 68.02 \\
\cline{2-10}
& Claude 3.5 Sonnet & 3.40 & 26.61 & 5.16 & 1.28×10$^9$ & -1.85 & 0.84 & 0.93 & 80.97 \\
\cline{2-10}
& GPT-4o & 3.73 & 29.42 & 5.42 & 3.44×10$^8$ & -3.27 & 0.82 & 0.92 & 74.80 \\
\hline
\end{tabular}
\label{tab:evaluation_metrics}
\end{center}
\end{table*}

\begin{figure}
    \centering
    \includegraphics[width=1\linewidth]{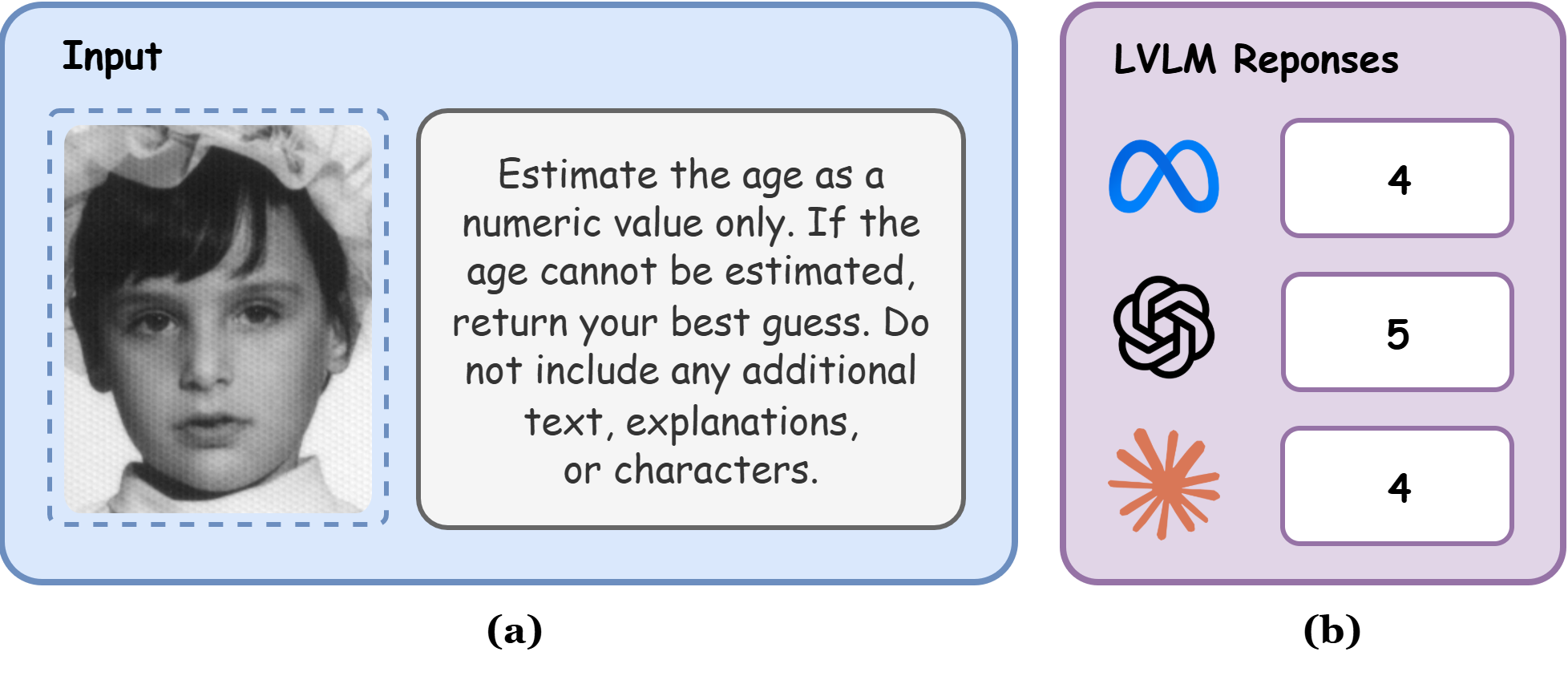}
    \caption{Zero-shot facial age estimation by LVLMs. (a) An input image and standardized prompt instruct the model to return a numeric age without explanation. (b) Predictions from LLaMA 3.2 Vision, GPT-4o, and Claude 3.5 Sonnet are shown, demonstrating their numeric outputs for the same input.}
    \label{fig:single-estimation}
\end{figure}

\begin{figure}[htbp]
\centerline{\includegraphics[width=\columnwidth]{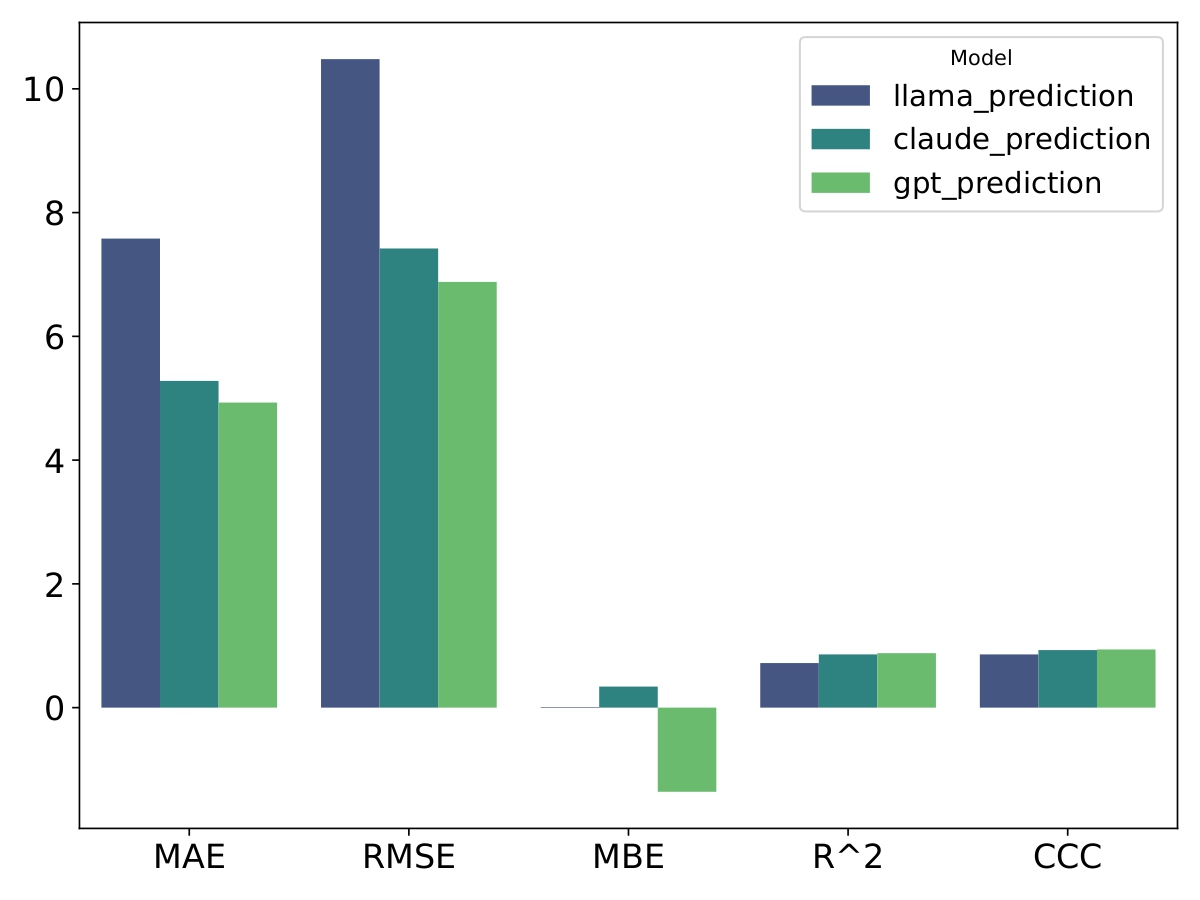}}
\caption{Model Performance Comparison on UTKFace Dataset}
\label{fig:comparison_utkface}
\end{figure}

\subsection{Performance Comparison}

Table~\ref{tab:evaluation_metrics} presents a comprehensive summary of evaluation metrics for different models across both datasets. The results reveal significant variations in performance between models and datasets.

\begin{figure} [H]
\centerline{\includegraphics[width=\columnwidth]{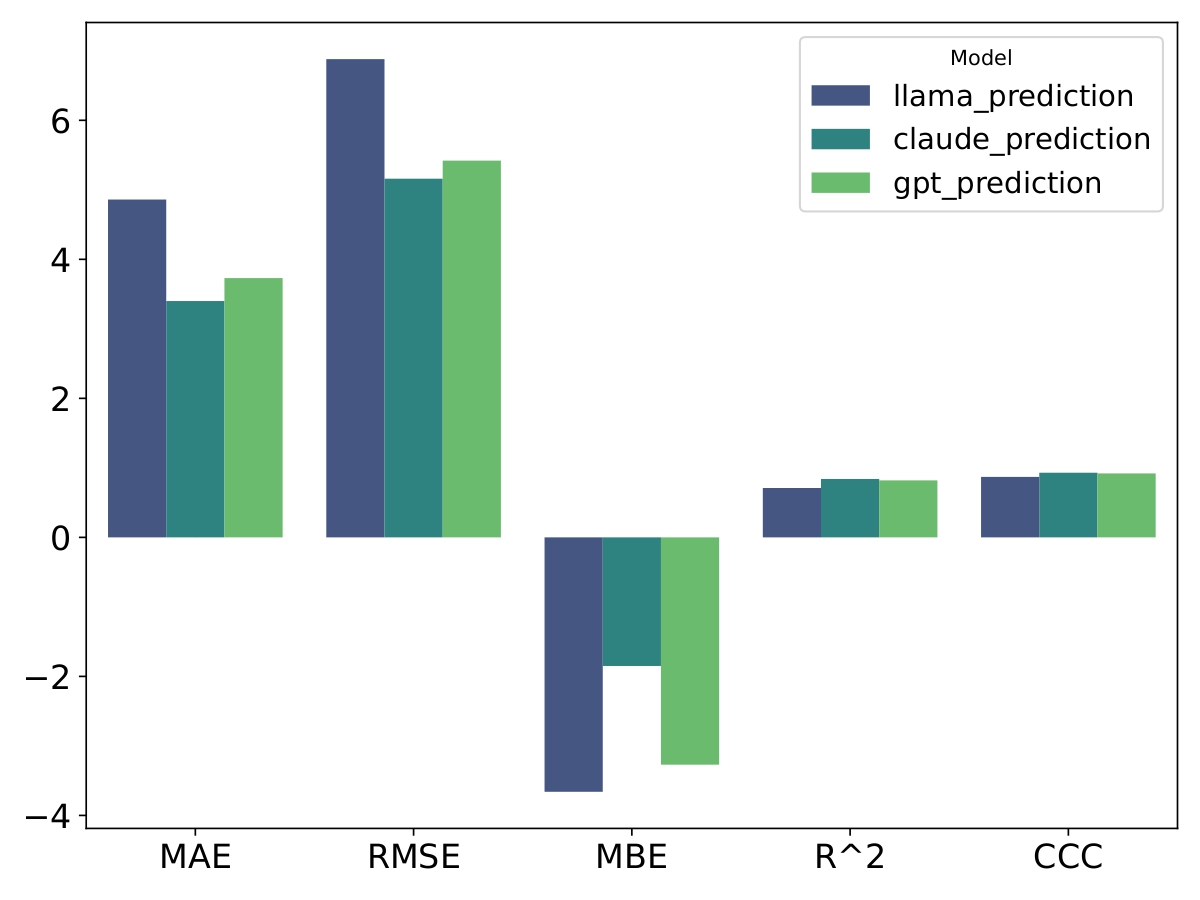}}
\caption{Model Performance Comparison on FG-NET Dataset}
\label{fig:comparison_fgnet}
\end{figure}

For the UTKFace dataset, which includes a wide age range (0-116 years) and diverse demographic groups, GPT-4o achieves the best performance with the lowest Mean Absolute Error (MAE) of 4.93 years, alongside the lowest Mean Squared Error (MSE) (47.35) and Root Mean Squared Error (RMSE) (6.88). This model also leads in terms of the highest R² (0.88) and Concordance Correlation Coefficient (CCC) (0.94), suggesting a strong ability to capture and predict age across the broad range of data. Claude 3.5 Sonnet performs competitively, with an MAE of 5.28 years and high CCC (0.93), but falls short in accuracy when compared to GPT-4o. LLaMA 3.2 Vision shows a relatively higher MAE (7.58) and lower R² (0.72), reflecting its challenges in generalizing across the diverse UTKFace dataset.

For the FG-NET dataset, which includes a narrower age range (0-69 years) and a smaller number of samples (1,002 images), Claude 3.5 Sonnet outperforms the other models with an MAE of 3.40 years and the highest accuracy within ±5 years at 80.97\%. GPT-4o follows closely with an MAE of 3.73 years and accuracy within ±5 years at 74.80\%, while LLaMA 3.2 Vision performs similarly to GPT-4o but is outperformed in both MAE (4.86) and R² (0.71).

The predictions from these models are visualized in Fig. \ref{fig:single-estimation}, where Fig. \ref{fig:single-estimation}(a) shows an input image and standardized prompt, instructing the models to output numeric ages. Fig. \ref{fig:single-estimation}(b) compares the predictions from LLaMA 3.2 Vision, GPT-4o, and Claude 3.5 Sonnet for the same input, further emphasizing the variation in their outputs.

\subsection{Model Performance Analysis}

GPT-4o consistently shows superior performance across both datasets. On UTKFace, it achieves the lowest MAE (4.93), MSE (47.35), and RMSE (6.88), indicating that it is the most accurate model in terms of absolute error. The high R² (0.88) and CCC (0.94) scores suggest that GPT-4o not only performs well in terms of precision but also captures the variability of age well, making it a robust model for age estimation. Claude 3.5 Sonnet offers strong performance, especially on FG-NET, where it excels with the lowest MAE (3.40) and the highest accuracy within ±5 years (80.97\%). While it is slightly less accurate than GPT-4o on UTKFace, it performs exceptionally well on the smaller FG-NET dataset, benefiting from the constrained age range.

LLaMA 3.2 Vision shows solid performance, but its higher MAE (7.58 on UTKFace and 4.86 on FG-NET) and lower R² (0.72 on UTKFace) suggest that it struggles more with the generalization task. This model seems to underperform compared to the commercial models, though it could still be useful for many practical applications, particularly in situations where computational resources are limited.
\begin{figure}
    \centering
    \includegraphics[width=1\linewidth]{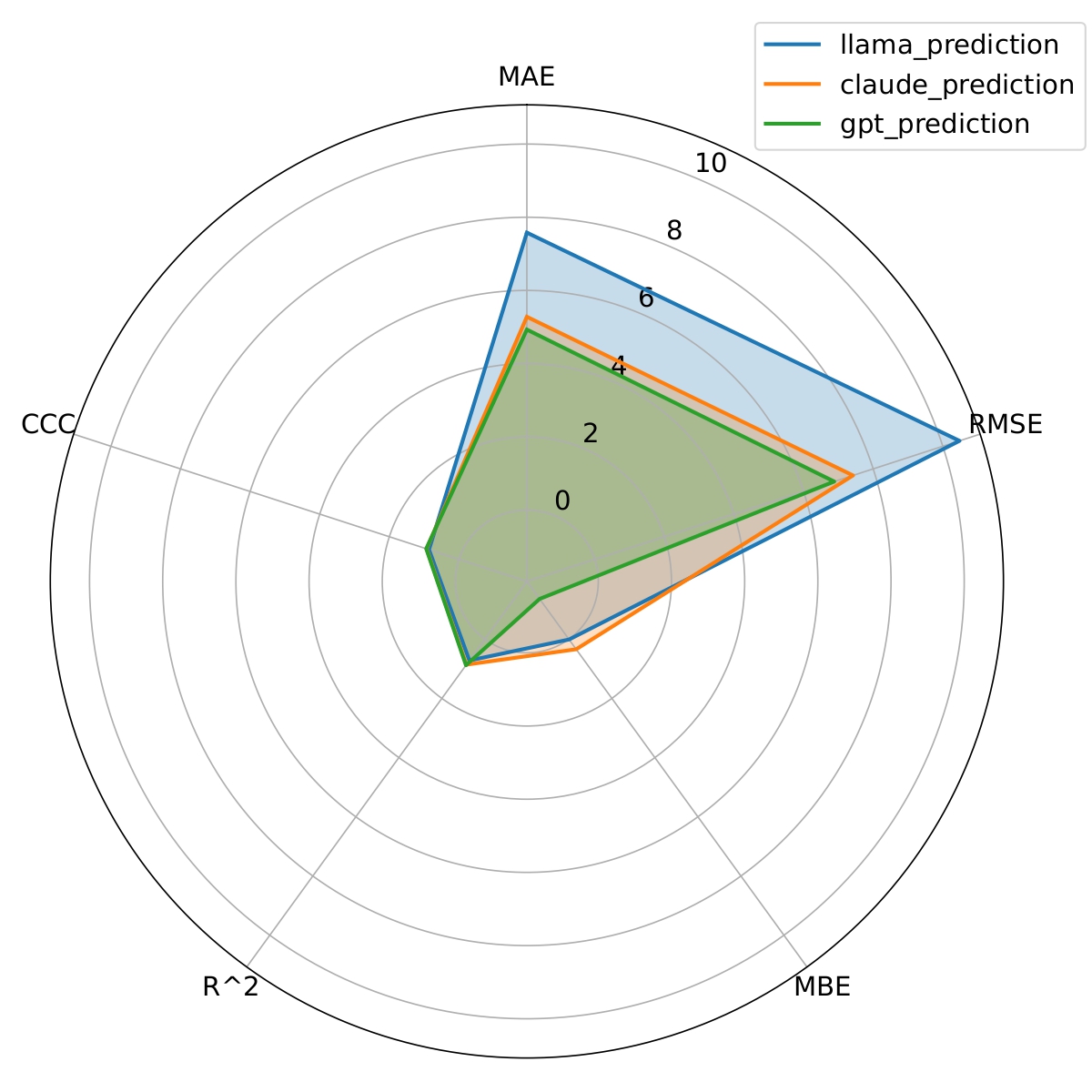}
    \caption{Radar chart visualization of model performance on UTKFace dataset, illustrating the relative advantages of each LVLM across different evaluation criteria.}
    \label{fig:radar-utkface}
\end{figure}

\subsection{Cross-Dataset Performance}

Across datasets, all models perform better on the FG-NET dataset compared to the UTKFace dataset. The smaller age range in FG-NET (0-69 years) means the models face less variability and therefore tend to produce more accurate predictions. In contrast, the broader age range in UTKFace (0-116 years) introduces more complexity, resulting in higher error rates.

An interesting finding from the MAPE values is the inflation of percentage errors when predicting ages of very young subjects, especially in FG-NET. This suggests that traditional error metrics like MAPE are not ideal for datasets with a substantial proportion of younger individuals, where small absolute errors can lead to disproportionately high percentage errors. Thus, it may be necessary to incorporate additional or alternative evaluation metrics for more robust model assessment, particularly for age estimation tasks involving young subjects.

\begin{figure}[H]
    \centering
    \includegraphics[width=1\linewidth]{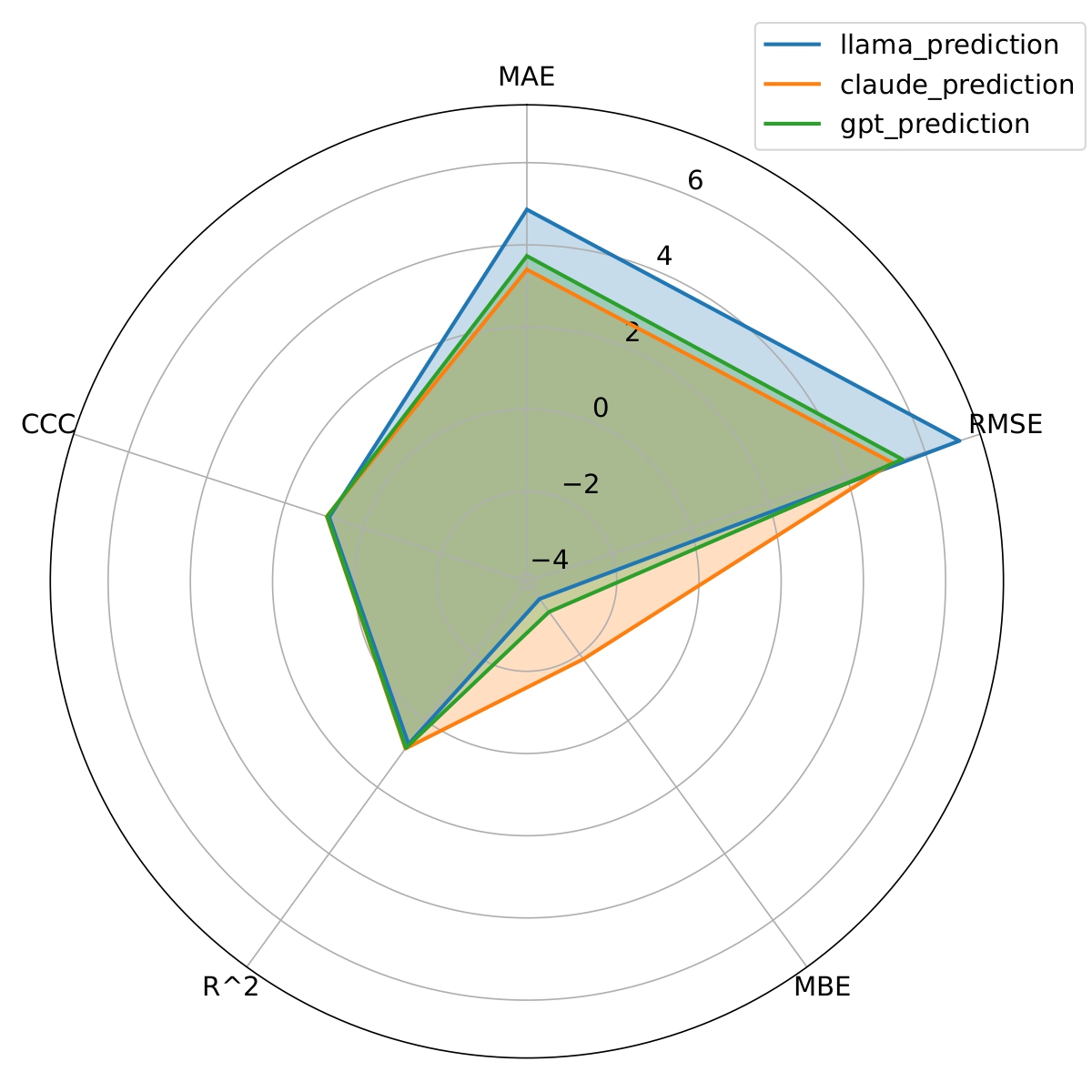}
    \caption{Radar chart visualization of model performance on FG-NET dataset. Each axis represents a different evaluation metric, providing intuitive comparison of model strengths and weaknesses.}
    \label{fig:radar-fgnet}
\end{figure}
The visualization in Fig.~\ref{fig:comparison_utkface} and Fig.~\ref{fig:comparison_fgnet} illustrate the relative performance of each model across different metrics for both datasets. Additionally, Fig.~\ref{fig:radar-utkface} and Fig.~\ref{fig:radar-fgnet} provide radar chart visualizations, highlighting model strengths and weaknesses across key performance metrics for the UTKFace and FG-NET datasets, respectively.

\section{Conclusions}

This paper provides the first comprehensive benchmark evaluation of large vision-language models (LVLMs) for zero-shot facial age estimation, establishing baseline performance metrics and uncovering valuable insights that could guide future advancements in multimodal biometric analysis. Due to the black-box nature of commercial LVLM APIs, we do not perform attention-based or saliency-based interpretability analysis in this work and leave it to future studies on open-weight models. Our experimental evaluation reveals that modern LVLMs possess remarkable emergent capabilities for age estimation without requiring task-specific training, marking a significant shift from traditional domain-specific approaches.

Among the models evaluated, GPT-4o consistently delivers the best performance across datasets, achieving the lowest mean absolute error (MAE) on UTKFace and competitive results on FG-NET. Its strong performance underscores the potential of general-purpose LVLMs in biometric applications. Claude 3.5 Sonnet excels particularly on the FG-NET dataset, indicating that it may have advantages when working with smaller, more constrained age ranges. On the other hand, LLaMA 3.2 Vision, despite being an open-source model, demonstrates promising results, though it generally lags behind the commercial models in accuracy and precision.

Overall, the LVLMs evaluated in this study show strong potential for real-world applications, with the best models achieving over 66\% accuracy within ±5 years on UTKFace and over 74\% on FG-NET. These results highlight the viability of LVLMs as practical tools for age estimation in diverse contexts.

However, our study also highlights the importance of selecting the right evaluation metrics. In particular, we found that MAPE can be misleading when evaluating datasets with a significant proportion of young subjects, due to the mathematical nature of percentage errors in such cases. Therefore, a more nuanced approach to model evaluation is needed to ensure robust and fair assessment of model performance.

Looking forward, future research should focus on fine-tuning LVLMs for specific biometric tasks, exploring multimodal fusion techniques that combine LVLMs with specialized biometric features, and conducting evaluations across more diverse demographic groups. In addition, bias-aware prompting and post-hoc demographic calibration could be explored to mitigate the performance disparities observed across subgroups. Future extensions of VLAgeBench will also consider few-shot and fine-tuned LVLM settings to directly compare zero-shot generalization with task-adapted performance. These efforts will help further improve the accuracy, fairness, and generalizability of age estimation models, ensuring that they can be used effectively in a wide range of real-world applications.



\bibliographystyle{IEEEtran}
\bibliography{./references}
\end{document}